\documentclass{article}

\usepackage[preprint]{neurips_2025}

\usepackage[utf8]{inputenc} 
\usepackage[T1]{fontenc}    
\usepackage{hyperref}       
\usepackage{url}            
\usepackage{booktabs}       
\usepackage{amsfonts}       
\usepackage{nicefrac}       
\usepackage{microtype}      
\usepackage{xcolor}         
\usepackage{csquotes}
\usepackage{graphicx}
\usepackage{amsmath}
\usepackage{mathtools}
\usepackage{bbold}
\usepackage{amsthm}

\newtheorem{definition}{Definition}

\newtheorem{proposition}{Proposition}
\newtheorem{lemma}{Lemma}
\newcommand{\overbar}[1]{\mkern 1.5mu\overline{\mkern-1.5mu#1\mkern-1.5mu}\mkern 1.5mu}

\title{ResNets Are Deeper Than You Think}

\author{%
	Christian H.X. Ali Mehmeti-Göpel \\
	Department of Computer Science\\
	Johannes-Gutenberg University\\
	Mainz, Germany\\
	\texttt{chalimeh@uni-mainz.de} \\
	\And
	 Michael Wand \\
	Department of Computer Science\\
	Johannes-Gutenberg University\\
	Mainz, Germany\\
	\texttt{wandm@uni-mainz.de} \\
}

\begin{document}

\maketitle

\begin{abstract}
	Residual connections remain ubiquitous in modern neural network architectures nearly a decade after their introduction.  Their widespread adoption is often credited to their dramatically improved trainability: residual networks train faster, more stably, and achieve higher accuracy than their feedforward counterparts. While numerous techniques, ranging from improved initialization to advanced learning rate schedules, have been proposed to close the performance gap between residual and feedforward networks, this gap has persisted. In this work, we propose an alternative explanation: residual networks do not merely reparameterize feedforward networks, but instead inhabit a different function space. We design a controlled post-training comparison to isolate generalization performance from trainability; we find that variable-depth architectures, similar to ResNets, consistently outperform fixed-depth networks, even when optimization is unlikely to make a difference. These results suggest that residual connections confer performance advantages \textit{beyond} optimization, pointing instead to a deeper inductive bias aligned with the structure of natural data.
\end{abstract}
\section{Introduction}

Combining deep neural networks and big data has enabled a series of dramatic breakthroughs; first in computer vision \citep{alexnet} but soon extending to a variety of different domains such as text processing \citep{llama3}, protein folding \citep{alphafold}, and multimodal tasks \citep{gemini}. The fact that this apparent success seems to extend to almost every type of natural data might be one of the most surprising results of modern deep learning research. In many cases, the inductive bias provided by network architectures and training protocols appears to be better aligned with natural data distributions than any other methods available today. This makes it at least plausible that we have found something akin to a universal prior for natural data, hidden within the inductive bias behind deep learning.

Over the course of the last decade, thousands of different network architectures and design patterns were proposed. Out of those, very few stood the test of time, in the sense that they are still commonly used years after their introduction. Additionally, it seems to be the case that when swapping one network component for another, similar performances are often reached, making it difficult to draw conclusions regarding their potential hidden inductive biases. Notable exceptions to this rule include Transformer blocks \citep{transformer}, normalization layers \citep{batchnorm}, and residual connections \citep{resnet}, which are still to this day ubiquitous in current architectures. In this work, we focus on the latter, aiming to take a deeper look at the reasons for the immense success and perseverance of the residual architecture design since its introduction. 

\cite{resnet} show that deep feedforward networks suffer from a {degradation problem}, where training performance deteriorates with depth, and propose residual connections as a \emph{reformulation}  that lets a
layer learn the identity map more easily. The term {reformulation}, however, is ambiguous: it could mean either (i)~a pure \emph{reparametrisation} of the same function class, or (ii)~the introduction of a genuinely different hypothesis space.
Follow‑up work explored both interpretations.
Mean‑field analysis by \citet{meanfield_residual} traces the degradation
to numerical instability: the authors show that well-conditioned \emph{q- and c-maps}, which guarantee a forward/backward pass and non-vanishing rank, are highly predictive for network trainability and that adding skip-connections mitigates these problems. Yet, \cite{deep_kernel_shaping} and \cite{tat} show that even after one eliminates such numerical pathologies through carefully shaped initialisation and re-scaling of the nonlinear layers, residual nets still outperform equally well‑conditioned plain nets, suggesting additional factors at play.

A complementary perspective is given by \cite{resnet_ensemble}, who unravel a ResNet into an {ensemble} of exponentially many paths of different lengths and hypothesize that since long paths result in very low gradient norms, they effectively might not contribute to the total gradient in any meaningful way. They conclude that residual connections do not magically make deep paths trainable; instead, they shorten the \emph{effective path length} and thereby avoid exploding/vanishing gradients. However, despite being centered on the architectural shape of residual networks, their work only discusses the implications for trainability properties. 

Despite extensive study, past literature still conveys an incomplete picture of the benefits provided by residual connections. We provide an analytical argument showing that adding skip-connections {alters} the network’s function space rather than being a simple re‑parametrization. After undergoing an extensive literature review showing the difficulty of disentangling trainability and generalization properties, we supply new experiments showing that \emph{variable‑depth} architectures (i.e. containing both long and short paths) such as ResNets outperform \emph{fixed‑depth} feed‑forward networks (i.e. only containing long paths), even in a setting where the differences in trainability are negligible. This evidence, in line with past findings, suggests that the function space induced by variable‑depth networks might be more closely aligned with real‑world data distributions than the space defined by their fixed‑depth counterparts, implying that the long-standing performance gap between them may never be fully closed.

\section{Motivation}
\label{sec:related_work}
\subsection{A Brief History of Numerical Issues in Neural Network Training}

When training deep neural networks, certain basic numerical constraints must be met to optimize a network effectively. In this section, we show a list of commonly used necessary criteria for trainability as proposed by \cite{beyond_bn} and \cite{shattered_gradients}, along with a short history of how researchers addressed those issues in the past. For the remainder of this paper, we will refer to these specific issues simply as ``trainability issues''.

\begin{enumerate}
	\item\textbf{Stable Forward Propagation.} The scale of a network's activations should not grow exponentially across layers during the forward pass. This effect was also sometimes called covariate shift (e.g. \citep{bn_covshift} Figure 2). Earlier network architectures, such as VGG  \citep{vgg} without normalization layers, had difficulties scaling beyond $\sim 20$ layers due to this issue.
	\item\textbf{Non-Shattering Gradients / Stable Backward Propagation.} 	Gradients with respect to network inputs should maintain some degree of spatial auto-correlation / smoothness (ref. \citep{shattered_gradients}, Fig. 1) for successfully training with SGD, particularly with momentum. \cite{ringing_relus} show that the classical ``exploding gradients'' problem in deep feedforward networks (ref. \citep{meanfieldbn}) can be seen as a consequence of shattering gradients, as the gradients wrt. the network's weights also shatter, which in turn leads to high average gradient norms. \cite{autotune_ours} (ref. Fig 5) and \cite{deep_information_propagation} show that exploding gradients can severely impede a network's trainability.
	
	\item\textbf{Informative Forward Propagation:} The output rank, or more generally the singular value spectrum of the last layer's activations of a network, should not collapse. When the activations of the last layer become linearly dependent, the effective dimensionality of the network function is reduced, and as a consequence, the network loses the ability to distinguish between different inputs and cannot learn effectively. This effect has been shown to occur for the product of many Gaussian matrices (ref. \cite{deep_linear} Figure 6) and also impedes a network's trainability \citep{deep_information_propagation}.
	
\end{enumerate}

The real difficulty lies in solving all of these problems at the same time, as a solution to a single problem can worsen the remaining ones, as we will see in the following.

\paragraph{Initialization Schemes} Researchers first attempted to solve the first two issues by utilizing cleverly designed initialization schemes that guarantee a stable forward and backward propagation  \citep{glorot_init}. Assuming that the effect of the nonlinear layers in the network is negligible, the authors find that for layers of different width, an initialization scheme can either satisfy condition 1 or 2 at the same time and derive an initialization scheme that average both approaches as a compromise; however, this approach breaks down for deeper networks of non-constant width. \cite{he_init} later extended this idea to account for the effect of the nonlinear layer used on activations and gradients.

\paragraph{Dynamic Isometry / DKS}There is a later line of research called \textit{dynamic isometry}, which aims to initialize the network in a way that brings all singular values of the input-output Jacobian of the network function close to 1, consequently solving all three stability issues. For linear networks at initialization, it is possible to achieve dynamic isometry for networks with arbitrary depth using i.i.d. orthogonal matrices \citep{dynamic_isometry_linear}, but not for i.i.d. Gaussian matrices. As for nonlinear networks with sigmoid activations, a similar result can be found \citep{dynamic_isometry_nonlinear}, but the proof strategy involves shrinking the pre-activations into the linear regime of the activation, effectively linearizing the network. \citep{deep_kernel_shaping} recognize this problem and find a solution to create truly nonlinear deep feedforward networks that are also numerically stable. However, their approach not only relies on weight initialization but additionally requires modifications of the network, mainly activation function transformations.

\paragraph{Normalization Layers} Normalization layers like Batch Normalization \citep{batchnorm} ensure a stable forward pass at initialization and throughout training as per construction. However, as these layers usually do not perform a full whitening transformation, the covariance between channels can still vanish. \cite{bn_rank_collapse} prove that BatchNorm retains at least the square root of the full rank at initialization for linear networks; however, their result does not hold for nonlinear networks or during training. There are attempts to implement full whitening normalization layers \citep{iternorm, decorrelated_bn} that also normalize covariances between channels, but these add significant computational overhead, guarantee spectral isotropy only in the forward pass but not in the backward pass, and bear several other issues such as stochastic axis swapping. Unfortunately, the combination of nonlinear layers and normalization layers also causes exploding gradients in the backward pass at initialization \citep{meanfield_bn, exploding_grads_blog}. 

\paragraph{Residual Connections} The further addition of residual connections ensures an informative forward pass along with a stable backward pass \cite{meanfield_bn}. When properly scaling the residual branch, a stable forward pass can be achieved even without normalization layers \citep{fixup}. Many works attempt to characterize the benefits gained by the addition of residual connections, but their argumentation often boils down to the compression of the singular spectrum across layers \citep{ntk_resnet, rethinking_resnet} or exploding/shattering gradients \citep{resnet_ensemble} mentioned above.

\subsection{Closing the Gap}
Researchers have attempted to close the performance gap between deep feedforward networks and ResNets originally described by \cite{resnet} for many years, but still fail to fully close it.  

\cite{autotune_ours} show that it is possible to eliminate exploding/shattering gradients even in arbitrarily deep normalized feedforward networks by warming-up the learning rate properly in early training, or simply normalizing the layer-wise gradient norms. However, even with these fixes, a significant performance gap is observable between deep feedforward and residual networks in Figure 5 of their work. 

\cite{meanfield_cnn} show that using a clever initialization scheme, it is possible to train 10,000-layer nonlinear convolutional neural networks. However, their construction involves practically linearizing the network at initialization and taking steps so small that the network remains mostly linear also during training; this effectively renders the network linear and drastically reduces its expressivity. The “looks-linear” initialization of \citep{shattered_gradients} follows a similar approach of initializing the network in a linear state, however, it suffers from trainability issues due to the nonlinear activations once training moves beyond the initial regime.

Instead of fully linearizing the network at initialization, another possible approach is to control the ``degree of nonlinearity'' of the nonlinear layers in the network: \cite{ringing_relus} (ref. Figure 9) show that the performance degradation of deep feedforward networks can be alleviated to a certain degree by using Leaky ReLU activations and controlling their slope. \cite{deep_kernel_shaping} follow a similar idea, but allow more flexibility, as in their work every nonlinear layer obtains a different scaling factor that is found by a solver. \cite{tat} further develop this idea and fix an issue of the approach specifically wrt. the ReLU nonlinearity and obtain even better results, but a generalization gap of over one percent on ImageNet remains for networks with 101 layers (ref. Table 9), even when trained with the expensive K-FAC \cite{kfac} optimizer. However, it is unclear whether optimizing numerically degenerate networks, such as deep feedforward networks, using a second-order optimizer, creates numerical instabilities in itself. Additionally, to compensate for the drastically different c- and q-maps of residual and non-residual architectures, their layer-wise scaling factors must be highly different; but these directly control the ``degree of nonlinearity'' of the networks, directly affecting their expressivity. Thus, this work does not yield a direct comparison of networks of similar nonlinear depth, which is the setup we are looking for.

We conclude that to our best knowledge, there have been no successful attempts to fully close the performance gap between deep feedforward and residual networks. As fixing the numerical instabilities of deep feedforward networks is a difficult problem, \textbf{it is unclear at this point whether the performance gap measured so far stems from side effects or newly introduced instabilities of proposed solutions such as DKS, or can be attributed to a an inductive bias \textit{beyond} trainability issues as defined above.
}
\subsection{Partial Linearization}
As a direct comparison of residual and feedforward networks in training is highly problematic for the reasons elucidated above, we opt for a different approach. Instead of opposing two architectures with potentially quite different trainability properties in a side-by-side training run from scratch, we attempt to extract a network with the relevant structure from an already trained network, therefore minimizing the impact of trainability properties. In this section, we take a look at existing frameworks that allow for such an undertaking.

\cite{nonlinear_advantage} and \cite{layer_folding} present similar techniques that reduce the amount of ReLU units in fully trained networks. Both works realize this in a post-training phase, where ReLU units are replaced with PReLU \citep{he_init} units and an additional regularization term, which simply penalizes every nonlinear unit in the network, is used to push their slopes towards linearity. The major difference is that the former authors do this at a channel granularity (i.e. one slope parameter per channel), whereas the latter use layer-wise units (i.e. a single slope parameter per layer). This difference, however, is crucial: by using a channel-wise technique, the resulting network can have variable-depth, akin to a ResNet; whereas when using the layer-wise technique, the resulting network has fixed depth, akin to a feedforward network. We use this difference to realize our comparison of the performance of variable- and fixed-depth networks. The authors have done plenty of comparisons between their methods, but the precise setup needed to make our claim (ref. Section \ref{sec:partial_lin}) is not covered.
\section{On Function Spaces}
\begin{displayquote}
	``We present a residual learning framework to ease the training of networks that are substantially deeper than those used previously. We explicitly reformulate the layers as learning residual functions with reference to the layer inputs, instead of learning unreferenced functions.''
\end{displayquote}
This quote from the abstract of \cite{resnet} explicitly describes ResNets as a ``reformulation'' of a non-residual network function. As this terminology is ambiguous, we first introduce precise definitions and notation.
\begin{definition}
	Let $f(x, \theta):\mathbb{R}^{n\times p}\rightarrow \mathbb{R}^m$ be a network function with flattened input vector $x \in \mathbb{R}^n$ and weight vector $\theta \in \mathbb{R}^p$. We then define a \textbf{reparametrization} of the network function $f$ as another network function $g(x, \theta):\mathbb{R}^{n\times p'}\rightarrow \mathbb{R}^m$ along with a weight reparametrization function $h(\theta):\mathbb{R}^{p'}\rightarrow \mathbb{R}^p$ such that:
	$$
	g(x, h(\theta)) = f(x, \theta).
	$$
	We call a reparametrization \textbf{equivalent}, if the networks $f$ and $g$ have the same width and depth.
\end{definition}

In Appendix Section \ref{sec:non_equivalent_reparam}, we first establish that in the general case with a non-injective nonlinearity and square matrices, it is not possible to equivalently reparametrize a ResNet as a feedforward network. Now we take a look at a simple constriction that shows that such a reparametrization is generally 
possible, albeit requires more parameters.

\begin{proposition}[Locally Linear Nonlinearity]
	\label{prop:general_nonlin}
	Let $R$ be a residual block defined as follows:
	\begin{equation}
		R(x) = \phi(\overbar{W}x + \overbar{b}) + x\\
	\end{equation}
	where $\overbar{W}\in\mathbb{R}^{n\times n}$ represents the weight matrix and $\overbar b\in\mathbb{R}^n$ the bias and $\phi:\mathbb{R}\rightarrow\mathbb{R}$ is an element-wise nonlinear function, which is differentiable in a single point $c\in\mathbb{R}$. 
	
	$R$ can then be reparametrized as a feedforward layer $F$ with one additional linear layer and double the width:
	\begin{equation}
			F(x) = W_2\phi(W_1 x + b_1) + b_2,
	\end{equation}
		with weights $W_1\in\mathbb{R}^{n\times 2n}$ $, W_2\in\mathbb{R}^{2n\times n}$ and biases $b_1\in\mathbb{R}^{2n}$, $b_2\in \mathbb{R}^n$ .
\end{proposition}
\begin{proof}
In a first step, we need to shift and shrink the input into the linear region of $\phi$ using a linear map  $L_1:\mathbb{R}^n\rightarrow\mathbb{R}^{2n}$. Let $\varepsilon>0$, we can then write:
\begin{equation}
	L_{1}(x)\;\coloneqq \; 
	\underbrace{\begin{bmatrix} \varepsilon \cdot I \\[2pt] \overbar W \end{bmatrix}}_{\displaystyle \mathbb R^{2n\times n }}
	x
	+
	\underbrace{\begin{bmatrix} c\cdot\mathbb{1}\\[2pt] \overbar b \end{bmatrix}}_{\displaystyle \mathbb R^{2n}}.
\end{equation}
Next, after applying the nonlinear layer $\phi$, we need to add, un-shrink, and shift back using  $L_2:\mathbb{R}^{2n}\rightarrow\mathbb{R}^n:$
\begin{equation}
	\label{eq:L2-def}
	L_{2}(x)\;\coloneqq\;
	\underbrace{\bigl[	\frac{1}{\varepsilon\,\phi'(c)}\cdot I\;\;I\bigr]}_{\displaystyle \mathbb R^{\,n\times 2n}}
	x
	\;-\;
	\underbrace{\frac{\phi(c)}{\varepsilon \phi'(c) }\cdot \mathbb{1}}_{\displaystyle \mathbb R^{\,n}}.
\end{equation}
We can now put together our feedforward layer $F=L_1\circ\phi\circ L_2$ and obtain:
\begin{equation}
	\label{eq:local_f2}
	F(x) = \frac{\phi(\varepsilon x + c ) }{\varepsilon \phi'(c)}+ \phi(\overbar{W}x+\overbar{b})
	\;-\;
	\frac{\phi(c)}{\varepsilon \phi'(c) }\cdot \mathbb{1}.
\end{equation}
If we now plug the Taylor expansion for the term we want to linearize
\begin{equation}
	\phi(\varepsilon x + c) = \phi(c)\cdot\mathbb{1} + \phi'(c) \varepsilon x + \mathcal{O}(\varepsilon^2 \lVert x\rVert ^2)
\end{equation}
into Equation \ref{eq:local_f2}, we finally obtain the reparametrization, which is exact for $\varepsilon\rightarrow0$ :
\begin{align}
	F(x) &= \frac{1}{\varepsilon \phi'(c)}(\phi(c)\cdot\mathbb{1} + \phi'(c) \varepsilon x + \mathcal{O}(\varepsilon^2 \lVert x\rVert ^2)+ \phi(\overbar{W}x+\overbar{b})
	\;-\;
	\frac{\phi(c)}{\varepsilon \phi'(c) }\cdot \mathbb{1}\\
	&= x  + \phi(\overbar{W}x+\overbar{b})+\mathcal{O}(\varepsilon \lVert x\rVert ^2),
\end{align}
\end{proof}

In Appendix Section \ref{sec:relu_reparam}, we show a similar construction specifically for $\phi=ReLU$ that works by shifting the data term into the linear region of the activation function instead of shrinking, thus not suffering from numerical issues related to the shrinking/unshrinking of the pre-activations.

\paragraph{Limitations}The construction above however requires significant assumptions:
\begin{itemize}
	\item Requires sufficient numerical precision in order mitigate numerical instabilities during shrinking/unshrinking of the pre-activations. Otherwise yields an error term if $\varepsilon$ is not chosen small enough.
	\item Requires boundedness of the pre-activations.
	\item Requires double the depth and width.
\end{itemize}
We don't see the halving of the number of layers as a major limitation, since we would typically compare networks with the same normalized average path length, i.e. a feedforward network with $\ell$ layers with a ResNet with approximately $\ell /2$ layers. Especially in normalized networks, the boundedness of the pre-actiations does not seem like a major constraint. However, the effective halving of the width of the network is likely to affect the performance significantly.

\textbf{In conclusion, it is impossible to reparametrize a residual network as an equivalent feedforward network and possible reparametrizations} (for a single block) \textbf{require additional width.}

\section{Empirical Evidence}
\begin{figure}[tbp]
	\centering
	\begin{minipage}[t]{0.495\linewidth}\vspace{0mm}%
		\centering
		\includegraphics[width=\linewidth]{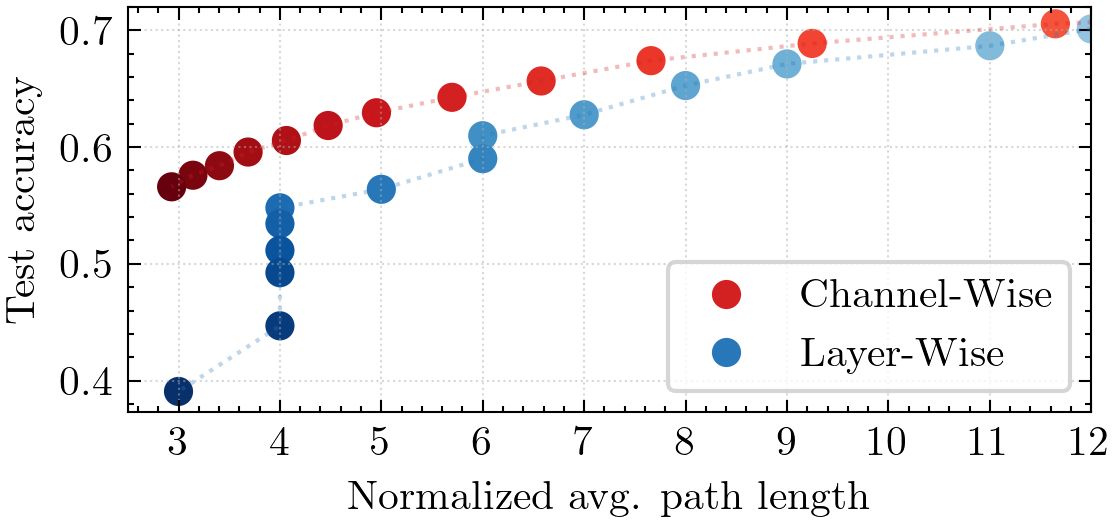}
	\end{minipage}
	\hfill
	\begin{minipage}[t]{0.495\linewidth}\vspace{0mm}%
		\centering
		\includegraphics[width=\linewidth]{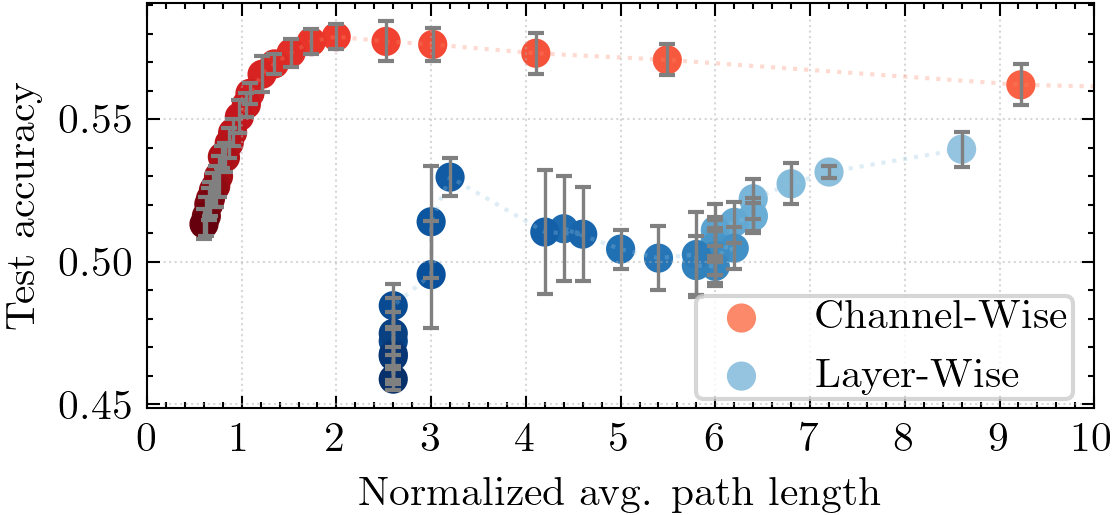}
	\end{minipage}
	\caption{Comparing the test accuracies of partially linearized networks using a channel-wise or layer-wise partial linearization approach on the \textbf{ImageNet (left)} and \textbf{Cifar100 (right)} datasets. The results of the latter are averaged over 5 runs per data point, and standard deviations are indicated as error bars.}
	\label{fig:partial_lin_imgnet_cifar100}
\end{figure}
\label{sec:partial_lin}
As we have now established that ResNets span a different function space than feedforward networks, it is now plausible that these networks outperform feedforward networks beyond trainability issues. As this question fundamentally depends on how well the network architecture's prior aligns with the training data, it is empirical in nature. As we established in the past section, numerical instability (i.e. an unstable forward/backward pass or vanishing rank in the last layer) is greatly impeding the trainability of deep feedforward networks, so we need to exert great care in our experimental design to avoid such issues. 

The basic idea is a setup vey similar to \cite{nonlinear_advantage, layer_folding}: we start from an already pre-trained network and then mold it into different shapes in a post-training phase, in order to reduce the impact of trainability on the comparison. Starting from a deep pre-trained feedforward network, we gradually reduce the number of nonlinear units in the network using an additional regularization term that penalizes each nonlinear unit in the network during a post-training phase. Our goal is to mold the network either into a variable-depth network or a fixed-depth network during the post-training phase in order to compare their respective performances. We achieve this by using channel-wise, respective layer-wise linearization approach in otherwise equal experimental settings: starting from a deep feedforward network, a channel-wise linearization can result in a variable-depth network (i.e. similar to a ResNet) if only a proportion of the channels within a layer become linear; a layer-wise linearization always results in a fixed-depth network (i.e. a shallower feedforward network). As the networks are molded only into their respective shapes only \textit{after} the network reaches its full performance, we do not expect trainability to differently affect both runs.

\subsection{Implementation Details}
In order to reduce the amount of nonlinear units in a trained network, we replace its ReLU units with channel/layerwise PReLU units \citep{he_init} and add a sparsity regularization term 
\begin{equation}
	L_{0.5}= \sum |1 - \alpha_i|^{0.5}
\end{equation}
to the regular training loss scaled with a \textit{regularization weight} $\omega$, where $\alpha_i$ is the variable slope of the i-th PReLU. This way, the networks are incentivized to regularize some of their nonlinear units towards linearity, while preserving performance as much as possible. We set and freeze the slope parameter to $\alpha_i=1$ if it gets close enough to one, i.e. $\lvert \alpha_i - 1 \rvert< 0.01$.

Similar to \cite{nonlinear_advantage}, we measure the depth of the resulting partially linearized networks in average path length, which represents the average amount of nonlinear units encountered in a path from input to output through the computation graph of the network. Particularly, we use the width-agnostic normalized average path length (NAPL) as a measure. For networks linearized layer-wise, the NAPL of the resulting network is simply its depth minus one.

We chose RepVGG \citep{repvgg} as our starting architecture, as it is a recent network architecture without skip-connections but with competitive ImageNet performance, with a reasonable amount of parameters, and publicly available weights. The RepVGG architecture does contain skip-connections across linear layers, which in our framework does not affect nonlinear depth here measured by NAPL. We specifically chose an architecture with residual connections across the linear layers to highlight that nonlinear depth, which is only influenced by residual connections across nonlinear layers, is decisive of a network's expressivity. We chose specifically the RepVGG-A2 architecture with 23 layers, as it has a low number of parameters but still achieves an ImageNet performance of $76.4\%$, similar to a ResNet50. The post-training phase has a duration of 10/60 Epochs on ImageNet resp. Cifar100. We chose a longer post-training duration on the Cifar100 dataset to show that this effect does not vanish when (post-)training close to convergence. More implementation details can be found in the Appendix Section \ref{sec:arch_details}.

\subsection{Experimental Results}
\label{sec:results_repvgg}

In Figure \ref{fig:partial_lin_imgnet_cifar100} (left), we show the test accuracies of the resulting partially linearized networks using a channel-wise and layer-wise technique starting from a pre-trained RepVGG-A2 model. We can clearly see that starting for networks with NAPL under 12, the layer-wise approach starts to be less performant than the channel-wise approach, and that the gap becomes bigger for shallower networks. In the Appendix Section \ref{sec:partial_lin_cifar}, we show similar results for the Cifar10 and Cifar100 datasets. Interestingly, the NAPL where the performances of the two approaches diverge seems to be lower on easier datasets.

We also repeat this experiment on the Cifar100 dataset 5 times with a longer post-training phase, and report the results with error bars in Figure \ref{fig:partial_lin_imgnet_cifar100} (right). Interestingly, we see a slight increase of performance towards NAPL of around 3; this is possible as with increasing linearization the loss surface is also smoothened and therefore generalization performance can in fact slightly increase as a result for intermediate $\omega$, before dropping for lower $\omega$ as a result of low expressivity. Also on the Cifar100 dataset, we observe a significant performance gap between the layer-wise and channel-wise variants for lower NAPL.

\subsection{Comparing to Linearized ResNets}
\begin{figure}
	\begin{minipage}[t]{0.495\linewidth}\vspace{0mm}%
		\centering
	\includegraphics[width=\linewidth]{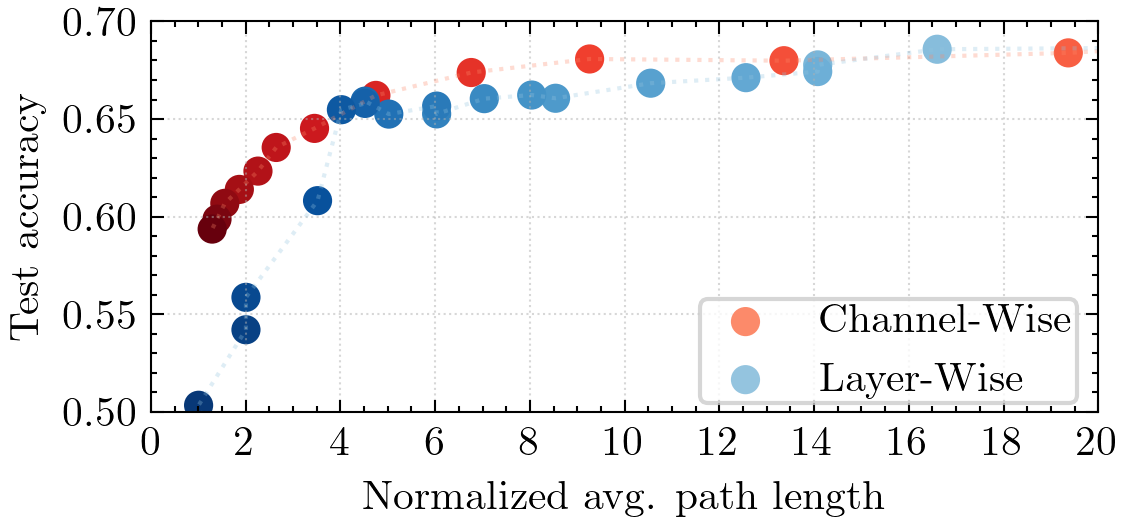}
	\end{minipage}
	\hfill
	\begin{minipage}[t]{0.495\linewidth}\vspace{0mm}%
		\centering
	\includegraphics[width=\linewidth]{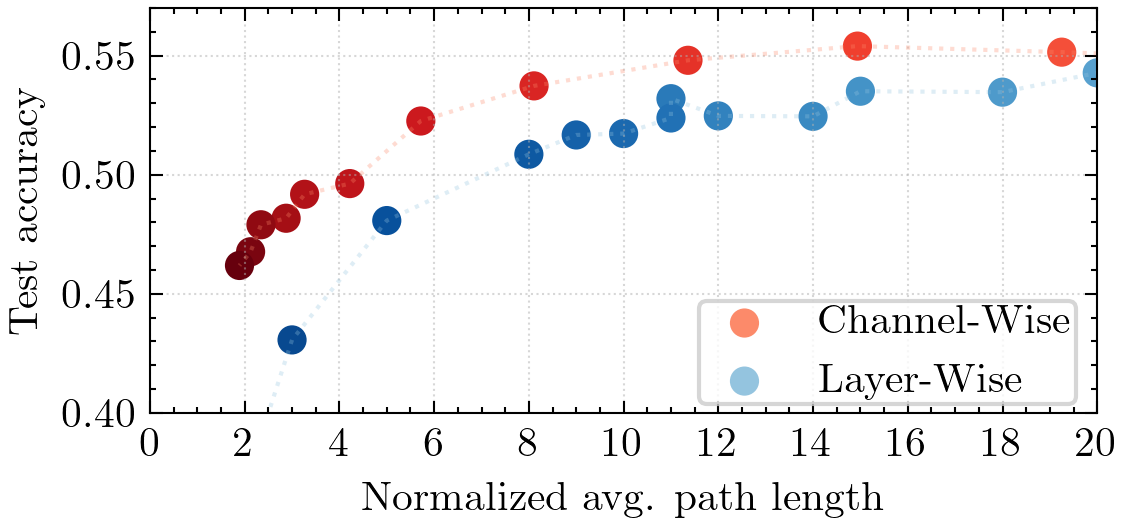}
	\end{minipage}
	\caption{Comparing the test accuracies on Cifar100 of partially linearized networks using a channel-wise or layer-wise partial linearization approach, stating from a \textbf{ResNet56 Short (left)} and \textbf{ResNet56 NoShort (right)}.}
	\label{fig:linearize_rn}
\end{figure}
\label{sec:results_rn}
In this Section, we repeat the results above, but start our linearization process from a pre-trained residual network instead of a pre-trained feedforward network. If the effect we saw in Section \ref{sec:partial_lin} is reduced, this imples that part of the observed benefit of the variable-depth path lengths can be replicated by regular residual connections. For a setup with comparable residual and feedforward networks, we use a ResNet56 ``Short'' (i.e. with residual connections) and ``NoShort'' (i.e. without residual connections). When linearizing a ResNet in Figure \ref{fig:linearize_rn} (left), we see that the layer-wise extracted networks only slightly underperform the channel-wise extracted networks for NAPL > 4. For NAPL 4 and below, we still see a bigger difference, which we conjecture is due to quantization artifacts, as there are only very few free slope parameters left in this case. The gap is significantly bigger when linearizing a feedforward network (right). 

\subsection{Shape of the Extracted Networks}
\label{sec:histo}
\begin{figure}
	\begin{minipage}[t]{0.495\linewidth}\vspace{0mm}%
	\centering
	\includegraphics[width=\linewidth]{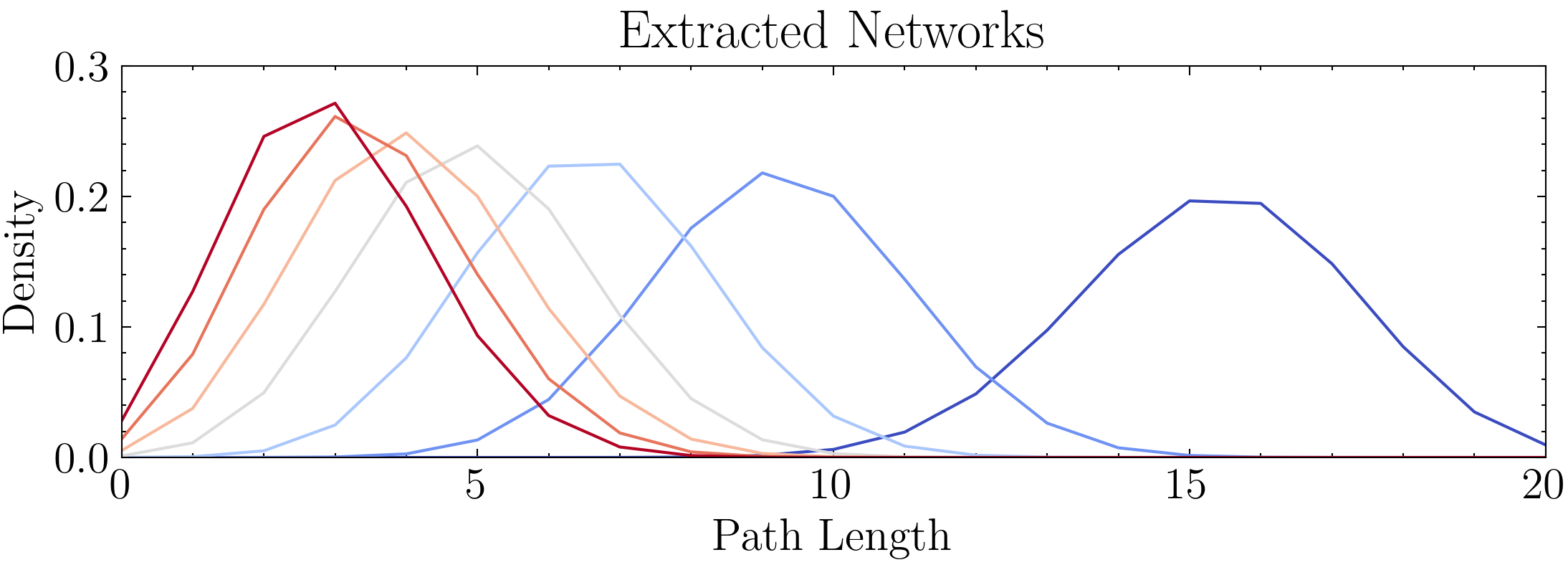}
\end{minipage}
	\hfill
	\begin{minipage}[t]{0.495\linewidth}\vspace{0mm}%
		\centering
	\includegraphics[width=\linewidth]{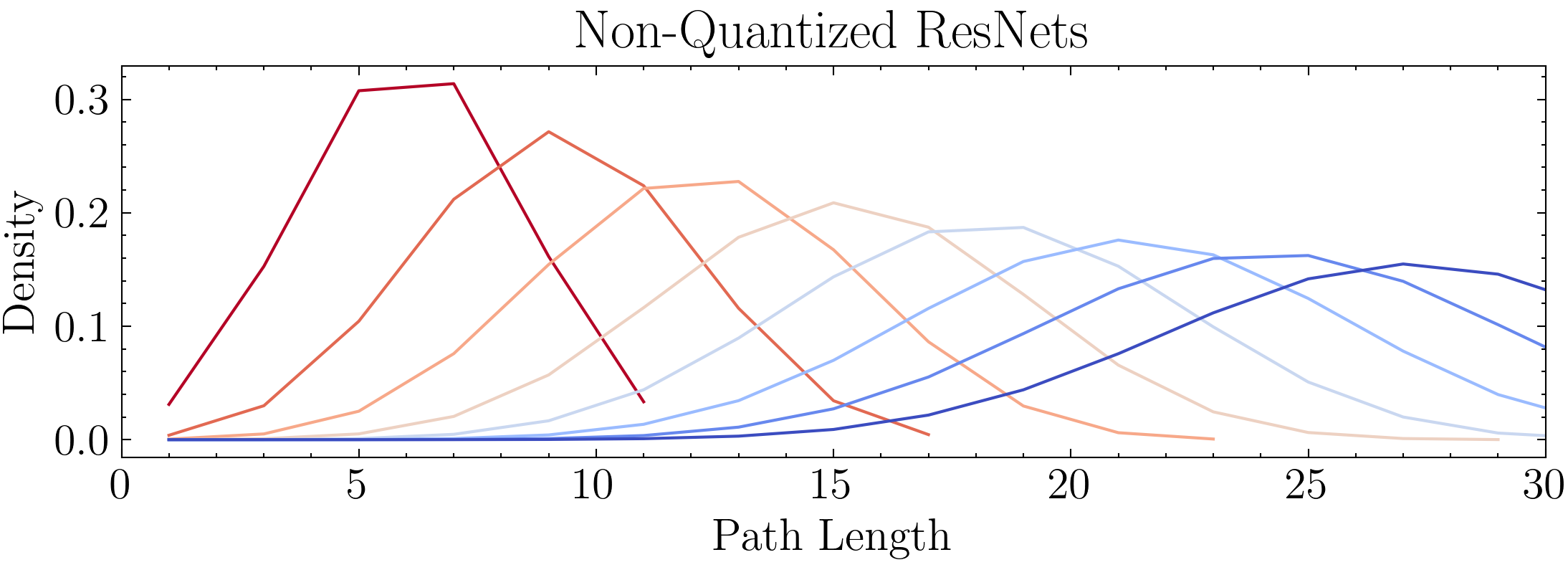}
	\end{minipage}
	\label{fig:hists}
	\caption{Comparing the histograms of networks extracted via\textbf{ partial linearization (left) }versus \textbf{standard ResNets (right)}. The red color indicates a lower $\omega$ (left) or a lower depth (right).}
\end{figure}

In this Section, we analyze the shape of the resulting partially linearized networks from Section \ref{sec:partial_lin} (ImageNet dataset) as a histogram of path lengths. For reference, we also include the histogram of path lengths for the non-quantized ResNets; we omit the zero-density points resulting from the block length 2 for better legibility. In Figure \ref{fig:hists} on the right-hand side, we see that the resulting distribution closely resembles a binomial distribution as predicted by \cite{resnet_ensemble}. Observed discrepancies from the theoretical binomial distribution in shallower models can be explained by architectural components beyond linear and PReLU layers, present in the actual network but excluded from the simplified theoretical model. In Figure \ref{fig:hists} (left), we observe that the extracted networks also contain a mixture of short and long paths, fairly similar to the standard ResNets. Note that this is not trivial, as these networks are derived from a fixed-depth model, and could very well also result in a fixed-depth network as well; their shape can be regarded as an emergent property of optimization.

\subsection{Limitations}
In this Section, we discuss the limitations of our experimental results. First, in the experiment from Section \ref{sec:partial_lin}, the partially linearized networks are extracted via gradient descent \textit{after} training, which is necessary to limit the impact of the drastically different trainability properties of variable-depth and fixed-depth networks on our experiment. There is no guarantee that networks trained from scratch behave the same way, although it appears likely. Also, since we are dealing with a non-convex optimization problem, which we optimize via a local gradient descent strategy, there is no guarantee that the networks extracted represent the exact sub-network with the best possible test accuracy given their target shape, i.e. that we reach a global minimum in both the channel-wise and layer-wise case. However, since merely the (relatively short) post-training phases differ between both approaches, and the network has already reached convergence before partial linearization, we argue that optimization is unlikely to make a difference in this comparison. Because of this limitation of our experimental setup, the results should be viewed as just another indication, along with a vast corpus of similar observations made in past literature (ref. Section \ref{sec:related_work}), that point to a similar conclusion.

Another limitation of this experiment is that the partially linearized network using the channel-wise approach has slightly more free parameters compared to the layer-wise approach. To be exact, the parameter difference represents 7786 out of $\sim26\cdot10^6$ parameters or $\sim 0.03\%$ of the total amount. We address this concern in the Appendix Section \ref{sec:missing_params}, where we present additional experiments that show that this small difference is highly unlikely to cause the performance gap observed.

Finally, one could argue that the performance gap we measured is a result of some other artifact of our experimental setup and cannot purely be attributed to the differences in the shape of the network. We addressed this issue in Section \ref{sec:results_rn} and see that except for very low NAPL values, the gap is mostly vanishes. This supports our claim that the performance gains measured indeed are a result of the variable-depth shape of the network.

\section{Discussion}
In this work, we investigate a potential inductive bias induced by the function space of variable-depth networks, i.e. networks composed of long and short paths. First, using a simple analytical argument, we establish that the function space described by residual networks truly is different from the one described by feedforward networks, given that we constrain the networks to have the same shape (i.e. width and depth), as one would in a realistic setting. We show that is not possible to find an equivalent reparametrization of a residual network as a feedforward network and that simple constructions that realize such reparametrizations involve doubling the network's width, along with other limitations. This makes it plausible that the performance of variable-depth and fixed-depth networks can differ due to factors beyond trainability.

Next, since there is ``no free lunch'' in machine learning, any claim of an architectural inductive bias must be verified on real data. However, after an extensive study of relevant literature, we come to understand that disentangling trainability from inductive bias is exceptionally challenging in this context, as deep feedforward networks suffer from many numerical issues during training, whereas comparable residual networks do not. Issues related to the uncontrolled growth of activation and gradient scales appear relatively tractable and can be addressed through normalization layers and specific warm-up strategies. The compression of the singular value spectrum of activations across layers (vanishing rank issue), however, seems more difficult to manage. As shown by \cite{ntk_resnet} (Figure 1), the inner product structure of inputs is affected from the very first layer, rendering even shallow networks unsuitable in a direct comparison by simple training. It still remains unclear whether the performance gap that persists in literature even after attempting to fix numerical issues in deep feedforward networks is truly beyond such issues.

For these reasons, in this work, we opt for a substantially different strategy in our experiments. Starting from a fully-trained deep feedforward network, we give the network the ability to turn some of its channels fully linear (channel-wise approach) and apply regularization pressure that punishes every nonlinear unit to the same degree. We observe in Section \ref{sec:histo} that the resulting emerging sub-networks contain a mixture of long and short paths, not unlike a standard ResNet. Then, we repeat the same procedure, but constrain the network to only keep paths of the same length (layer-wise approach). Finally, we then compare the resulting generalization performance of the networks extracted using both approaches, where we match networks of the same nonlinear depth, which we quantify as the average amount of nonlinear units encountered on a path through the computation graph of the network (NAPL). We observe a significant performance gap between the extracted variable-depth architectures and their fixed-depth counterparts, even when controlling for average depth and parameter count, as we push the network’s nonlinear capacity toward the lower end. We further observe that this gap mostly disappears when the initial architecture is a ResNet, confirming that it arises from the constraint of allowing only long paths. We interpret this as further evidence that variable-depth networks outperform fixed-depth networks on natural data \textit{beyond} mere trainability. This finding aligns with trends observed in prior work and strengthens the case that variable-depth architectures can offer a genuine inductive advantage over fixed-depth networks.

\section{Conclusion}
Researchers attempted to close the performance gap between feedforward and residual networks for many years, yet no approach has fully succeeded. In this work, we reverse the question and ask: is it even possible, even under optimal training conditions? We conjecture that networks that are composed of a mixture of long and short paths permit an inductive bias that is well-aligned with natural data, which can result in performance benefits beyond trainability. Despite the difficulty of separating trainability issues and inductive bias in an experimental setting, we carefully design an experiment that extracts a sub-network of the relevant shape \textit{after} training and show a performance gap between fixed-depth and variable-depth networks.

\bibliographystyle{plainnat} 
\bibliography{references}

\appendix
\newpage
\section{Theory Appendix}
\label{sec:theory_appendix}
\subsection{Proof of Non-Equivalent Reparametrization}
\label{sec:non_equivalent_reparam}
\begin{lemma}
	Let $R:\mathbb{R}^n\rightarrow\mathbb{R}^n$ and $L:\mathbb{R}^n\rightarrow\mathbb{R}^n$ be a residual resp. feedforward block defined as follows:
	\begin{align}
			R(x) &= \phi(\overbar{W}x + \overbar{b}) + x\\
		F(x) &= \phi(Wx + b),
		\end{align}
	where $W,\overbar{W}\in\mathbb{R}^{n\times n}$ represent the weight matrices and $b,\overbar b\in\mathbb{R}^n$ the biases and $\phi:\mathbb{R}\rightarrow\mathbb{R}$ an element-wise non-injective nonlinear function. 
	
	We denote the set of all functions of these types as:
	
	\begin{align}
		\mathcal{R}(x)&\coloneqq \{R(\overbar W,\overbar b) | \overbar W\in\mathbb{R}^{n\times n},\overbar b\in\mathbb{R}^n \}\\
		\mathcal{F}(x)&\coloneqq \{F(W,b) |W\in\mathbb{R}^{n\times n},b\in\mathbb{R}^n \}.
	\end{align}
	
	Then the function space $\mathcal F$ and $\mathcal R$ differ, i.e.  $\mathcal F\neq \mathcal R$.
\end{lemma}
\begin{proof}
	For $W=b=0$, we obtain $R=I$, where $I$ is the identity function. We now show that $F$ is not injective, which implies that $I\not\in\mathcal{F}$.
	
	\textbf{Case 1:} $W$ is not invertible. As $W$ is square, $F$ is therefore not injective.\\
	\textbf{Case 2:} $W$ is invertible. We assumed that $\phi$ is not injective and thus there exist scalars $c_1,c_2\in\mathbb{R}$ with $x_1\neq x_2$ such that $\phi(x_1) = \phi(x_2)$. Since W is invertible, we can find vectors $x_1,x_2\in\mathbb{R}^n$ with $x_1\neq x_2$ such that $\phi(Wx_1+b)=\phi(Wx_2+b)$ and thus $F$ is not injective.
	
	If $F$ is not injective, $I\not\in F$.
\end{proof}

We now extend this result to entire networks, i.e. concatenation of such blocks.

\begin{proposition}
	We define the function space of residual resp. feedforward networks as: 
	\begin{align}
		\hat{\mathcal{G}}(x)&\coloneqq \{G_1(\overbar W_1,\overbar b_1) \circ \ldots \circ G_\ell (\overbar W_\ell,\overbar b_\ell)| \overbar W_\ell \in\mathbb{R}^{n\times n},\overbar b_\ell \in\mathbb{R}^n \quad \forall i=1\ldots\ell\}\\
		\hat{\mathcal{F}}(x)&\coloneqq \{F_1( W_1, b_1) \circ \ldots \circ F_\ell ( W_\ell, b_\ell)|  W_\ell \in\mathbb{R}^{n\times n}, b_\ell \in\mathbb{R}^n\quad \forall i=1\ldots\ell\}.
	\end{align}
		Then the function space $\hat{\mathcal{F}}$ and $\hat{\mathcal{R}}$ differ, i.e.  $\hat{\mathcal{F}} \neq \hat{\mathcal{R}}$.
\end{proposition}
\begin{proof}
	If we apply Lemma 1 to the first layer, the proof follows immediately.
\end{proof}
\subsection{Possible Reparametrizations}
\label{sec:relu_reparam}
In this Section, we show that in a network with ReLU nonlinearity, we can reparametrize a residual block as a feedfoward block using slightly different assumptions as in the construction in Proposition \ref{prop:general_nonlin}

\begin{proposition}[ReLU Nonlinearity]
	Let $R$ be a residual block defined as follows:
	\begin{equation}
		R(x) = \phi(\overbar{W}x + \overbar{b}) + x\\
	\end{equation}
	where $\overbar{W}\in\mathbb{R}^{n\times n}$ represents the weight matrix and $\overbar b\in\mathbb{R}^n$ the bias and $\phi:\mathbb{R}\rightarrow\mathbb{R}$ is the ReLU function.
	
	$R$ can then be reparametrized by a feedforward layer $F$ with one additional linear layer with double the width:
	\begin{equation}
		F(x) = W_2\phi(W_1 x + b_1) + b_2,
	\end{equation}
	
	with weights $W_1\in\mathbb{R}^{n\times 2n}$ $, W_2\in\mathbb{R}^{2n\times n}$ and biases $b_1\in\mathbb{R}^{2n}$, $b_2\in \mathbb{R}^n$ .
\end{proposition}

\begin{proof}
We attempt to construct our feedforward block $F$:

\begin{equation}
	F:x
	\;\xrightarrow{L_1}
	\begin{bmatrix}x+b_1\\ \overbar Wx+\overbar{b}\end{bmatrix}
	\;\xrightarrow{\phi}
	\begin{bmatrix}\phi(x+ b_1)\\ \phi(\overbar Wx+ \overbar b)\end{bmatrix}
	\;\xrightarrow{L_2}
	\phi(x+b_1) + \phi(\overbar Wx + \overbar{b}) + b_2,
\end{equation}

where $L_1:\mathbb{R}^n\rightarrow\mathbb{R}^{2n}$ is a linear map :

\begin{equation}
	L_{1}(x)\;\coloneqq \;
	\underbrace{\begin{bmatrix} I \\[2pt] \overbar W \end{bmatrix}}_{\displaystyle \mathbb R^{2n\times n}}
	x
	+
	\underbrace{\begin{bmatrix} b_1 \\[2pt] \overbar{b} \end{bmatrix}}_{\displaystyle \mathbb R^{2n}},
\end{equation}

where $I\in\mathbb{R}^{n\times n}$ denotes the identity matrix. Let $L_2:\mathbb{R}^{2n}\rightarrow\mathbb{R}^n$ be a second linear map: 

\begin{equation}
	L_2(x)\coloneqq \underbrace{\bigl[\;I\;\;I\bigr]}_{\mathbb R^{n\times 2n}}
	x+ \underbrace{b_2}_{\mathbb R^{n}}.
\end{equation}

We now have to make a significant assumption: the pre-activations are lower bounded by a constant, i.e. $\exists c\in\mathbb{R}$ : $x_i > c ×\quad \forall i \in \{1,\ldots,n\}.$ We can now set the biases $b_1,b_2\in \mathbb{R}^n$ as follows:

\begin{align}
	b_1 & \coloneqq c\cdot \mathbb{1} \quad \\
	b_2 & \coloneqq -c \cdot \mathbb{1},
\end{align}

where $\mathbb{1}\in\mathbb{R}^n$ denotes the all‑ones vector. This way, we shift the pre-activations out of the nonlinear range of the activation function and back, and obtain: 

\begin{equation}
	\phi(x+b_1) + b_2=x.
\end{equation}

As we now obtain $F(x) = R(x)$, we have found a reparametrization.
\end{proof}

This construction avoids numerical issues due to shrinking/unshrinking of the activations present in Proposition \ref{prop:general_nonlin}, but requires a bias term to work (necessary for shifting the pre-activations) which is not present in a standard ResNet, for example.
\section{Experimental Appendix}
In this Section, we present complementary experimental results to the main paper.

\subsection{Partial Linearization on Cifar 10}
\label{sec:partial_lin_cifar}
We repeat the experiments from Section \ref{sec:partial_lin} on the Cifar10 and show the results in Figure \ref{fig:partial_lin_cifar10}. We observe similar results to the ImageNet results discussed in the main paper, however, the NAPL, where the performance of the two techniques approximately breaks even, seems to be lower than for the ImageNet dataset: at around 4 NAPL. It appears that the harder the dataset, the higher the break-even point is.

\subsection{Partial Linearization with Different Measures}
\begin{figure}[tbp]
	\centering
	\begin{minipage}[t]{0.48\linewidth}\vspace{0mm}%
		\centering
		\includegraphics[width=\linewidth]{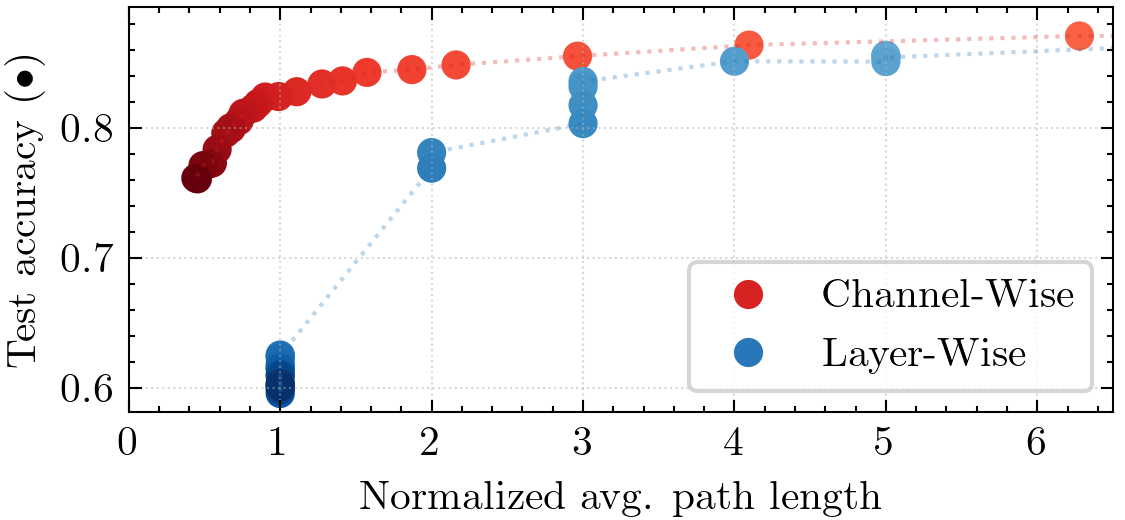}
		\caption{Comparing the test accuracies on \textbf{Cifar10} of partially linearized networks using a channel-wise or layer-wise partial linearization approach.}
		\label{fig:partial_lin_cifar10}
	\end{minipage}
	\hfill
	\begin{minipage}[t]{0.48\linewidth}\vspace{0mm}%
		\centering
		\includegraphics[width=\linewidth]{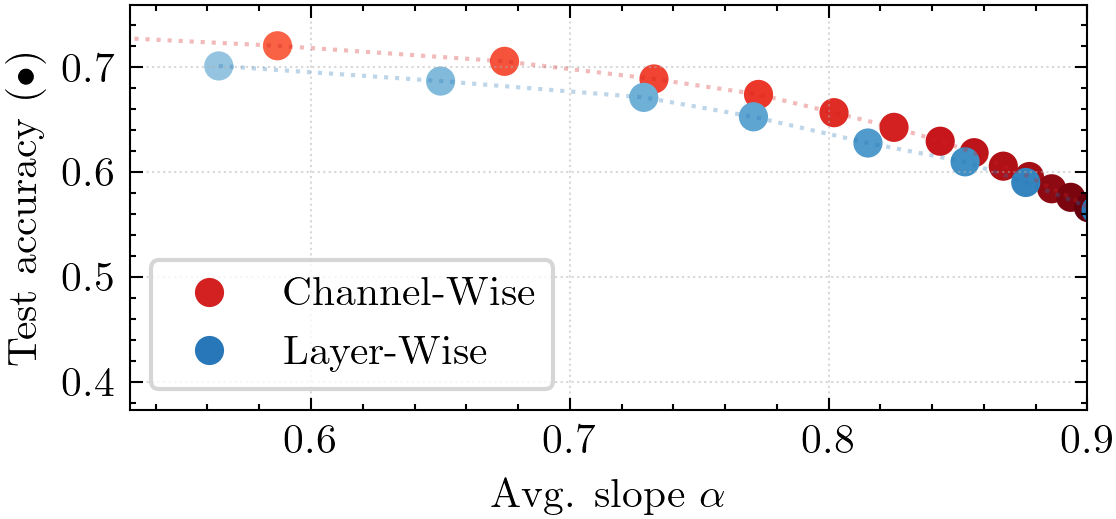}
		\caption{Comparing the test accuracies on \textbf{ImageNet} of partially linearized networks using a channel-wise or layer-wise partial linearization approach, using the  \textbf{average slope} measure.}
		\label{fig:partial_lin_imgnet_slope}
	\end{minipage}
\end{figure}

\begin{figure}[tbp]
	\centering
	\begin{minipage}[t]{0.48\linewidth}\vspace{0mm}%
		\centering
		\includegraphics[width=\linewidth]{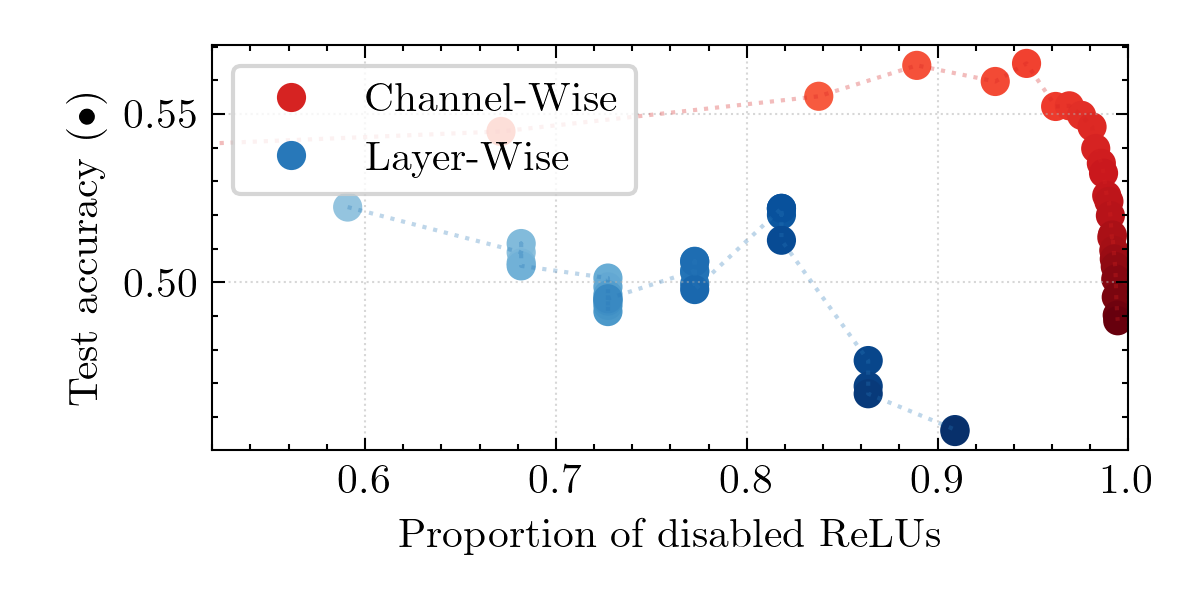}
		\caption{Comparing the test accuracies on \textbf{Cifar100} of partially linearized networks using a channel-wise or layer-wise partial linearization approach, using the  \textbf{percentage of disabled units} measure.}
		\label{fig:partial_lin_cifar100_pct}
	\end{minipage}
	\hfill
	\begin{minipage}[t]{0.48\linewidth}\vspace{0mm}%
		\centering
		\includegraphics[width=\linewidth]{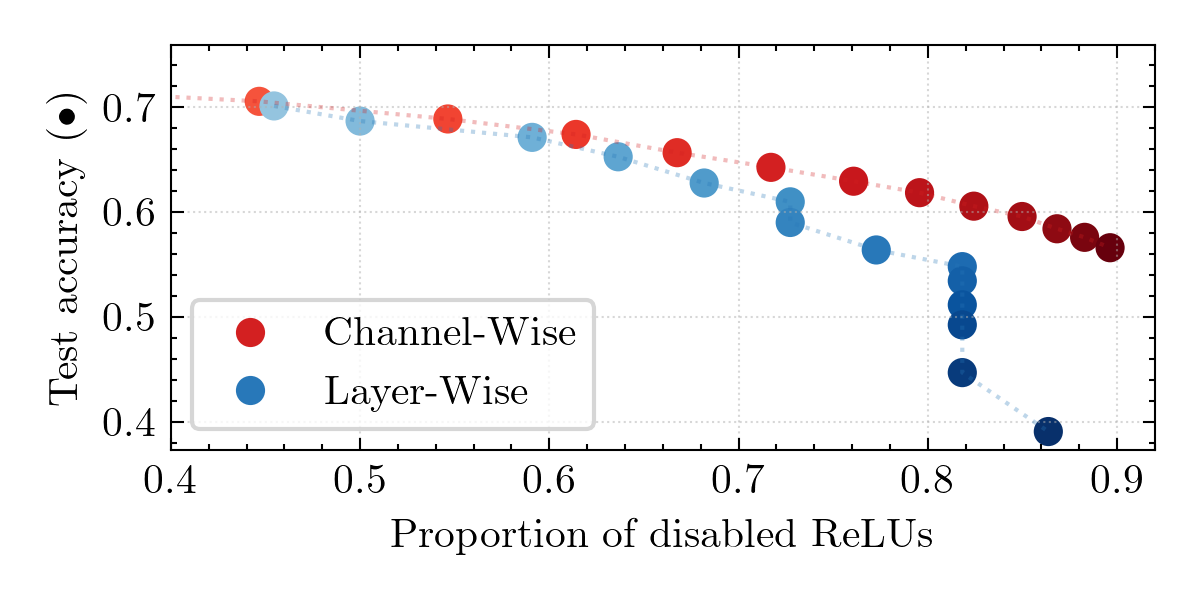}
		\caption{Comparing the test accuracies on \textbf{ImageNet} of partially linearized networks using a channel-wise or layer-wise partial linearization approach, using the  \textbf{proportion of disabled units} measure.}
		\label{fig:partial_lin_imgnet_pct}
	\end{minipage}
\end{figure}
In this Section, we replicate the experiment from Figure \ref{fig:partial_lin_imgnet_cifar100}, but we plot the average slope $\alpha=\frac{1}{n}\sum_{i=1}^{n}\alpha_i$ instead of the NAPL measure. We do this to confirm that the performance gap we observed cannot be attributed to a discretization artifact, as for the NAPL measure, a single channel is either fully linear or nonlinear. In Figure \ref{fig:partial_lin_imgnet_slope}, we see that even using the average slope as a measure of nonlinearity, we can observe the same phenomenon, as the networks linearized using the layer-wise approach have consistently lower performances compared to the ones linearized using the channel-wise approach.

In Figure \ref{fig:partial_lin_imgnet_pct} and \ref{fig:partial_lin_cifar100_pct}, we recreate the experiments of Figure \ref{fig:partial_lin_imgnet_cifar100} but show the results with the proportion of disabled PReLU units instead of the NAPL measure to confirm that the measured gap is not an artifact of this measure. As we can still observe a significant gap, this does not appear to be the case.

\subsection{Accounting for Missing Parameters}
\label{sec:missing_params}
In this Section, we address the concern that the results from Section \ref{sec:partial_lin} could be biased, since the networks extracted via the channel-wise method contain slightly more parameters (one slope parameter for each channel in a layer) than the networks extracted via the layer-wise method  (one slope parameter per layer). 

Sadly, this problem is not easily fixable via construction, as for example adding a linear layer before or after the nonlinearity would not be equivalent, but adding another non-regularized channel-wise PReLU unit before the layer-wise unit would add more nonlinear layers and thus void the experiments.

\subsection{Post-Post-Training}
\begin{figure}[tbp]
	\centering
			\begin{minipage}[t]{0.48\linewidth}\vspace{0mm}%

\includegraphics[width=\linewidth]{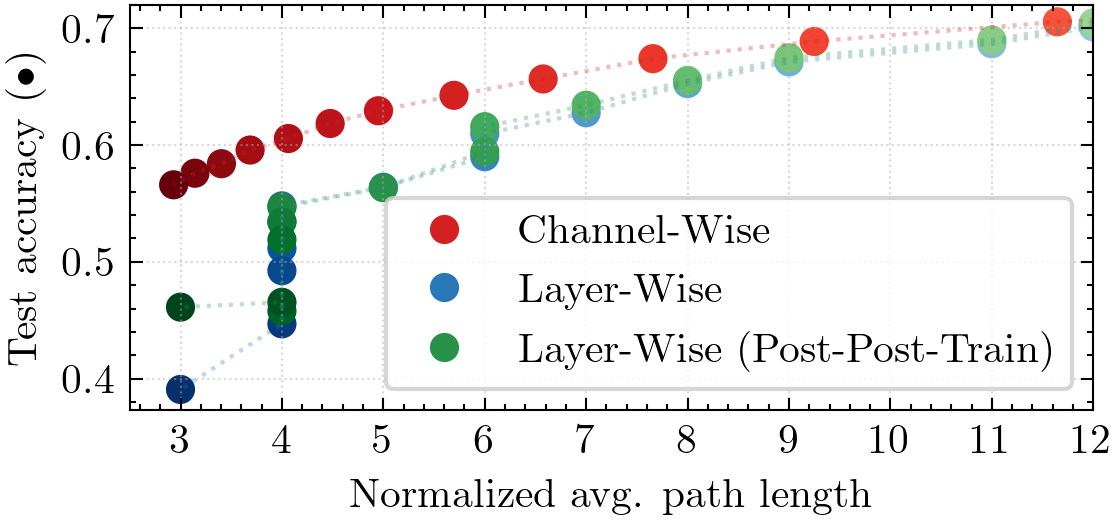}
\caption{Comparing the test accuracies on \textbf{ImageNet} of partially linearized networks using a channel-wise or layer-wise partial linearization approach, including ``post-post-training'' runs, where we replace any non fully linear layer-wise PReLUs with unregularized channel-wise PReLUs and train for another 10 epochs to account for missing parameters.}
\label{fig:partial_lin_imgnet_napl_postpostrain}
	\end{minipage}
	\hfill
	\begin{minipage}[t]{0.48\linewidth}\vspace{0mm}%
	\centering
\includegraphics[width=\linewidth]{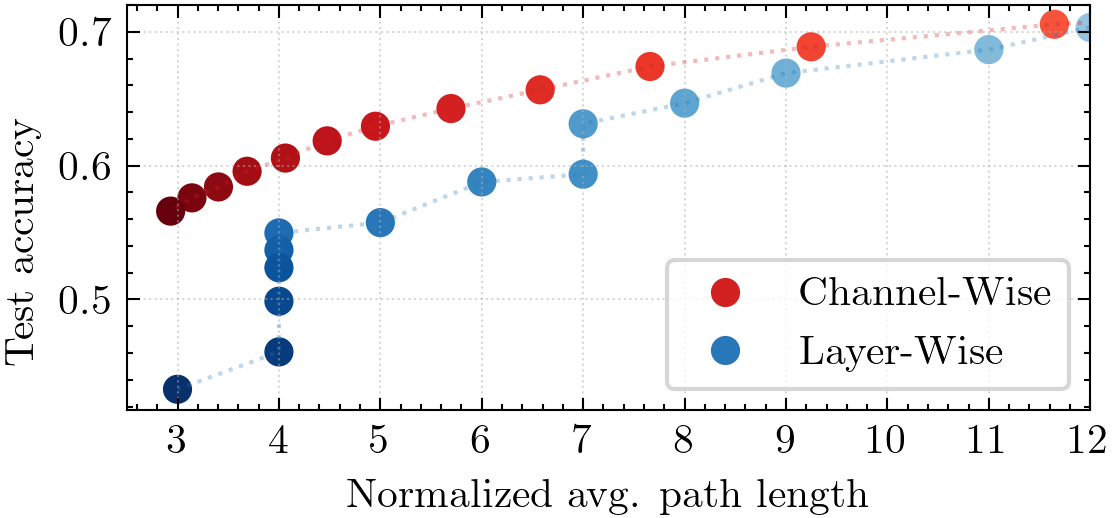}
\caption{Comparing the test accuracies on \textbf{ImageNet} of partially linearized networks using a channel-wise or layer-wise partial linearization approach, with an additional \textbf{channel-wise multiplication layer} after each nonlinearity to account for missing parameters in the layer-wise case.}
\label{fig:partial_lin_imgnet_napl_cwmult}
	\end{minipage}

\end{figure}

Starting from the post-trained networks in the layer-wise case, we replace all layer-wise PReLU units that are not fully linear ($\alpha_i$ < 1) by channel-wise units (with the existing slope as initialization parameter) and train again for 10 epochs, as in the original post-training, but without the regularization term. In Figure \ref{fig:partial_lin_imgnet_napl_postpostrain} we see that this yields just very slightly higher values and does not significantly reduce the effect strength.

\subsection{Channel-Wise Multiplier}
We add a channel-wise multiplication layer {after} the layer-wise PReLU layer to account for missing parameters and repeat the runs for the layer-wise linearization. Comparing Figure \ref{fig:partial_lin_imgnet_napl_cwmult} to Figure \ref{fig:partial_lin_imgnet_cifar100}(left), we do not observe a significant reduction in the performance gap between the two types of extracted sub-network.

\section{Hardware and Training Details}
\label{sec:arch_details}

The experiments in the paper were made on computers running Arch Linux, Python 3.13.3, PyTorch Version 2.7.0. The GPUs used were NVIDIA GeForce GTX 1080 Ti, GeForce RTX 2080 Ti and  GeForce RTX 4080 Ti. This post-training runs from Section \ref{sec:partial_lin} (on the ImageNet dataset) need approximately 2 days on a Nvidia GTX 4090.

The hyper-parameters in Figure \ref{fig:train_table} were used to (post-) train on the CIFAR-10 and CIFAR-100 reach a base test-accuracy of approximately $87.5\%$ for on CIFAR-10 and $54.6\%$ on Cifar100. As for the ImageNet runs, we used a pre-trained model from the RepVGG repository. For post-training, the hyper-parameters in Figure \ref{posttrain_table} were used.

\begin{figure*}[tb]
	\raggedright
	\begin{minipage}[t]{0.475\linewidth}\vspace{0mm}%
		\begin{tabular}{ l  l }
			\toprule\\
			Training& CIFAR-10 / CIFAR-100\\
			\midrule\\
			Epochs& 200\\
			Scheduler&Multistep ($\gamma=0.1$)\\
			Milestones&100, 150\\
			Learning rate&0.1\\
			Batch size&256\\
			Optimizer& SGD + Momentum\\
			Momentum&0.9\\
			Weight decay& 0.0001\\
			Augmentation& Random Flip\\
			\bottomrule
		\end{tabular}
	\end{minipage}
		\caption{Details of the training regime.}
	\label{fig:train_table}
\end{figure*}

\begin{figure*}[]
	\begin{minipage}[t]{0.475\linewidth}\vspace{0mm}%
		\begin{tabular}{ l  l }
			\toprule\\
			Post-Training& CIFAR-10 / CIFAR-100\\
			\midrule\\
			Epochs& 60\\
			Scheduler&Multistep ($\gamma=0.1$)\\
			Milestones&40, 50 (10, 22 for CINIC-10)\\
			Learning rate&0.01\\
			Batch size&256\\
			Optimizer& SGD + Momentum\\
			Momentum&0.9\\
			Weight decay& 0.0001\\
			Augmentation& Random Flip\\
			\bottomrule
		\end{tabular}
	\end{minipage}
	\hfill
	\begin{minipage}[t]{0.475\linewidth}\vspace{0mm}%
		\begin{tabular}{ l  l }
			\toprule\\
			Post-Training& ImageNet\\
			\midrule\\
			Epochs& 10\\
			Scheduler&Multistep ($\gamma=0.1$)\\
			Milestones& 6, 8\\
			Learning rate&0.01\\
			Batch size&256\\
			Optimizer& SGD + Momentum\\
			Momentum&0.9\\
			Weight decay& 0.0001\\
			Augmentation& Center Crop (224 px.)\\
			\bottomrule
		\end{tabular}
	\end{minipage}
	\caption{Details of the post-training regime.}
	\label{posttrain_table}
\end{figure*}

\newpage
\section*{NeurIPS Paper Checklist}

\begin{enumerate}
	
	\item {\bf Claims}
	\item[] Question: Do the main claims made in the abstract and introduction accurately reflect the paper's contributions and scope?
	\item[] Answer: \answerYes{} 
	\item[] Justification: We claim that ResNets span a different function space than feedforward network which presents benefits beyond trainability. We present constructive proof for the former and experimental evidence for the latter.
	
	\item {\bf Limitations}
	\item[] Question: Does the paper discuss the limitations of the work performed by the authors?
	\item[] Answer: \answerYes{} 
	\item[] Justification: We discuss the limitations explicitely in a separate sections
	
	\item {\bf Theory assumptions and proofs}
	\item[] Question: For each theoretical result, does the paper provide the full set of assumptions and a complete (and correct) proof?
	\item[] Answer: \answerYes{}
	\item[] Justification: Our proofs are complete and include assumptions made
	
	\item {\bf Experimental result reproducibility}
	\item[] Question: Does the paper fully disclose all the information needed to reproduce the main experimental results of the paper to the extent that it affects the main claims and/or conclusions of the paper (regardless of whether the code and data are provided or not)?
	\item[] Answer: \answerYes{}
	\item[] Justification: The full experimental setup is detailed including hyper-parameters.

	\item {\bf Open access to data and code}
	\item[] Question: Does the paper provide open access to the data and code, with sufficient instructions to faithfully reproduce the main experimental results, as described in supplemental material?
	\item[] Answer \answerYes{}
	\item[] Justification: Full code including launching scripts are provided. The datasets used are public available.

	\item {\bf Experimental setting/details}
	\item[] Question: Does the paper specify all the training and test details (e.g., data splits, hyperparameters, how they were chosen, type of optimizer, etc.) necessary to understand the results?
	\item[] Answer: \answerYes{},
	\item[] Justification: The paper contains a section that details all hyperparameters abd giw tget were chosen.
	
	\item {\bf Experiment statistical significance}
	\item[] Question: Does the paper report error bars suitably and correctly defined or other appropriate information about the statistical significance of the experiments?
	\item[] Answer: \answerYes{}
	\item[] Justification: Our main result is a rangetest for many values of $\omega$, and the trend is clear and significant. We added error bars in the Cifar100 case, as this is computationally more feasable as for the ImageNet runs.
	
	\item {\bf Experiments compute resources}
	\item[] Question: For each experiment, does the paper provide sufficient information on the computer resources (type of compute workers, memory, time of execution) needed to reproduce the experiments?
	\item[] Answer: \answerYes{}
	\item[] Justification: We reported the runtime in the experimental settings section.
	
	\item {\bf Code of ethics}
	\item[] Question: Does the research conducted in the paper conform, in every respect, with the NeurIPS Code of Ethics \url{https://neurips.cc/public/EthicsGuidelines}?
	\item[] Answer: \answerYes{}
	\item[] Justification: As this paper is about fundamental research, no particular ethical concerns arise.

	\item {\bf Broader impacts}
	\item[] Question: Does the paper discuss both potential positive societal impacts and negative societal impacts of the work performed?
	\item[] Answer: \answerNA{}
	\item[] Justification: This paper presents highly theoretical and foundational work whose goal is to advance the field of Machine Learning. There are many potential societal consequences of the rapid advancement of the field in general, but we deem this discussion out of the scope of this paper.
	
	\item {\bf Safeguards}
	\item[] Question: Does the paper describe safeguards that have been put in place for responsible release of data or models that have a high risk for misuse (e.g., pretrained language models, image generators, or scraped datasets)?
	\item[] Answer: \answerNA{}
	\item[] Justification: As highly theoretical and foundational work, this does not apply to our paper.
	
	\item {\bf Licenses for existing assets}
	\item[] Question: Are the creators or original owners of assets (e.g., code, data, models), used in the paper, properly credited and are the license and terms of use explicitly mentioned and properly respected?
	\item[] Answer: \answerYes{}
	\item[] Justification: We credit all relevant models and datasets used. Our code credits and respects licences.
	
	\item {\bf New assets}
	\item[] Question: Are new assets introduced in the paper well documented and is the documentation provided alongside the assets?
	\item[] Answer: \answerNA{}
	\item[] Justification: The paper does not introduce new assets.
	
	\item {\bf Crowdsourcing and research with human subjects}
	\item[] Question: For crowdsourcing experiments and research with human subjects, does the paper include the full text of instructions given to participants and screenshots, if applicable, as well as details about compensation (if any)? 
	\item[] Answer: \answerNA{}
	\item[] Justification: No human subjects were used.
	
	\item {\bf Institutional review board (IRB) approvals or equivalent for research with human subjects}
	\item[] Question: Does the paper describe potential risks incurred by study participants, whether such risks were disclosed to the subjects, and whether Institutional Review Board (IRB) approvals (or an equivalent approval/review based on the requirements of your country or institution) were obtained?
	\item[] Answer: \answerNA{}
	\item[] Justification: : No human subjects were used.
	
	\item {\bf Declaration of LLM usage}
	\item[] Question: Does the paper describe the usage of LLMs if it is an important, original, or non-standard component of the core methods in this research? Note that if the LLM is used only for writing, editing, or formatting purposes and does not impact the core methodology, scientific rigorousness, or originality of the research, declaration is not required.
	\item[] Answer:\answerNA{}
	\item[] Justification: This paper does not involve LLMs.
	
\end{enumerate}
\end{document}